\def\cvprPaperID{0000}
\def\confName{CVPR}
\def\confYear{2023}

\def\paperTitle{Enhancing 3D Transformer Segmentation Model for Medical Image with Token-level Representation Learning}

\def\authorBlock{
    Xinrong Hu$^{1}$ \qquad
    Dewen Zeng$^{1}$ \qquad
    Yawen Wu$^{1}$ \qquad
    Xueyang Li$^{1}$ \qquad
    Yiyu Shi$^{1}$ \\
    $^{1}$ Department of Computer Science and Engineering, University of Notre Dame \\
    {\tt\small \{xhu7, dzeng2, xli34, yshi4\}@nd.edu, yaw66@pitt.edu}
}

\newif\ifreview 
\newif\ifarxiv \newcommand{\arxiv}{\arxivtrue}
\newif\ifcamera 
\newif\ifrebuttal 

\arxiv 

\pdfoutput=1
\documentclass[10pt,twocolumn,letterpaper]{article}
\ifreview \usepackage[review]{cvpr} \fi
\ifarxiv \usepackage[pagenumbers]{cvpr} \fi
\ifrebuttal \usepackage[rebuttal]{cvpr} \fi
\ifcamera \usepackage{cvpr} \fi

\usepackage{graphicx}
\usepackage{amsmath}
\usepackage{amssymb}
\usepackage{booktabs}

\usepackage{times}
\usepackage{microtype}
\usepackage{epsfig}
\usepackage[table,xcdraw]{xcolor}
\usepackage{caption}
\usepackage{float}
\usepackage{placeins}
\usepackage{color, colortbl}
\usepackage{stfloats}
\usepackage{enumitem}
\usepackage{tabularx}
\usepackage{xstring}
\usepackage{multirow}
\usepackage{xspace}
\usepackage{url}
\usepackage{subcaption}
\usepackage{xcolor}
\usepackage[hang,flushmargin]{footmisc}

\ifcamera \usepackage[accsupp]{axessibility} \fi

% ...

\ifarxiv  \fi

\newcommand{\R}[1]{{%
    \textbf{%
        \ifstrequal{#1}{1}{\textcolor{red}{R#1}}{%
        \ifstrequal{#1}{2}{\textcolor{blue}{R#1}}{%
        \ifstrequal{#1}{3}{\textcolor{magenta}{R#1}}{%
        \ifstrequal{#1}{4}{\textcolor{teal}{R#1}}{%
                           \textcolor{cyan}{R#1}%
        }}}}%
    }%
}}

\usepackage{xr-hyper}

\makeatletter
\newcommand*{\addFileDependency}[1]{
  \typeout{(#1)}
  \@addtofilelist{#1}
  \IfFileExists{#1}{}{\typeout{No file #1.}}
}

\makeatother

\usepackage[pagebackref,breaklinks,colorlinks]{hyperref}
\usepackage[capitalize]{cleveref}
\crefname{section}{Sec.}{Secs.}
\crefname{table}{Table}{Tables}
\crefname{figure}{Fig.}{Figs.}

\begin{document}
\title{\paperTitle}
\author{\authorBlock}
\maketitle
\begin{abstract}
In the field of medical images, although various works find Swin Transformer has promising effectiveness on pixelwise dense prediction,  whether pre-training these models without using extra dataset can further boost the performance for the downstream semantic segmentation remains unexplored.
Applications of previous representation learning methods are hindered by the limited number of 3D volumes and high computational cost.
In addition, most of pretext tasks designed specifically for Transformer are not applicable to hierarchical structure of Swin Transformer.
Thus, this work proposes a token-level representation learning loss that maximizes agreement between token embeddings from different augmented views individually instead of volume-level global features.
Moreover, we identify a potential representation collapse exclusively caused by this new loss. 
To prevent collapse,  we invent a simple yet effective "rotate-and-restore" mechanism,
which rotates and flips one augmented view of input volume, and later restores the order of tokens in the feature maps. 
We also modify the contrastive loss to address the discrimination between tokens at the same position but from different volumes. 
We test our pre-training scheme on two public medical segmentation datasets, and the results on the downstream segmentation task
show more improvement of our methods than other state-of-the-art pre-trainig methods.
The codes are available at \href{https://github.com/xhu248/SimTROT}{https://github.com/xhu248/SimTROT}.
\end{abstract}

\section{Introduction}
\label{sec:introduction}

\begin{figure}
    \centering
    \includegraphics[width=0.8\linewidth]{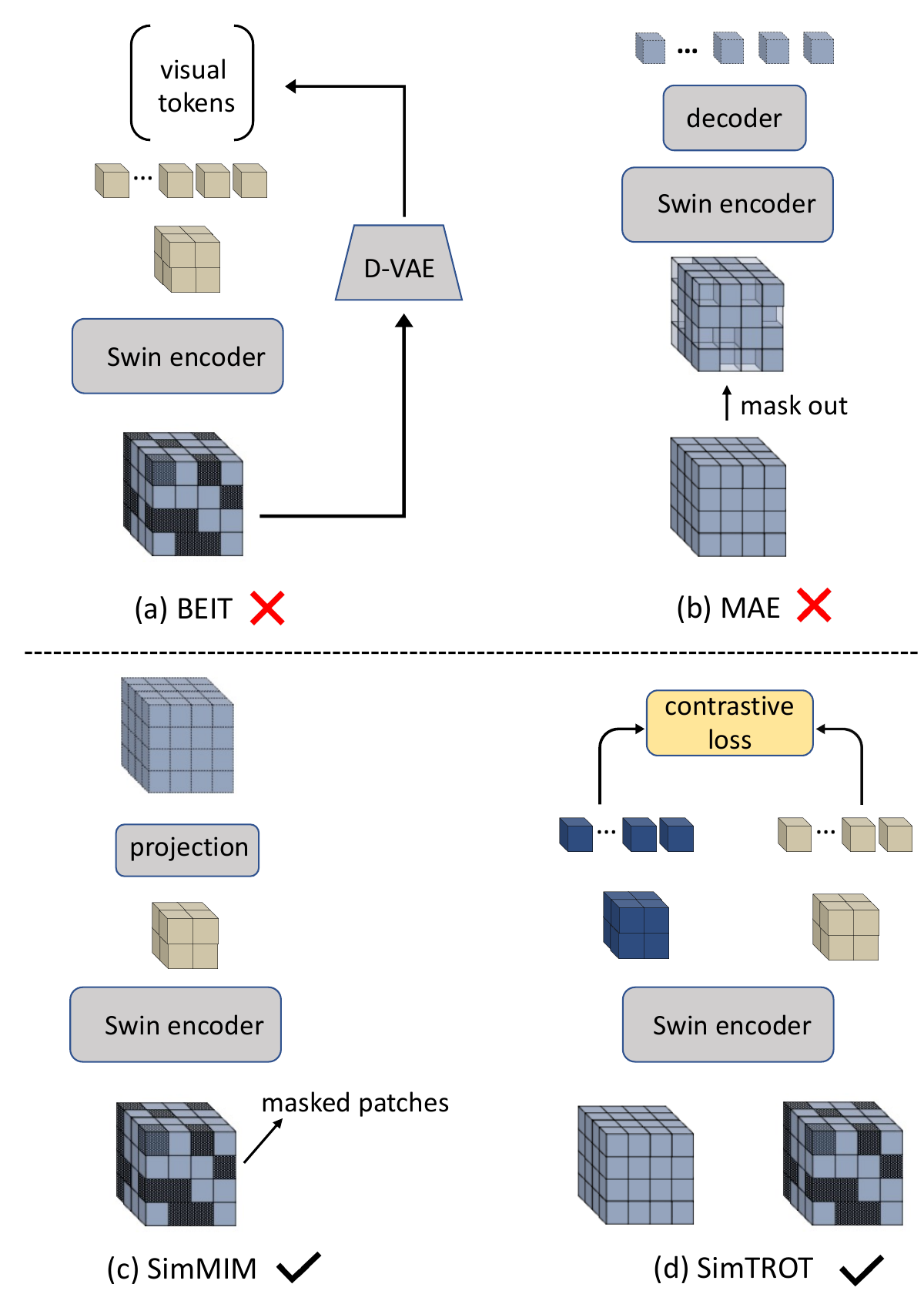}
    \caption{Illustration of three popular pre-training methods for Transformer and our SimPROT when the encoder backbone is changed to Swin Transformer. "\textcolor{red}{$\times$}" means the method is not feasible, while "\checkmark" means it is applicable. (a) BEIT\cite{bao2021beit} has unmatched number of patches between output and visual tokens, and for (b) MAE\cite{he2022masked}, volumes after masking out are not valid input for Swin Transformer. Notice that (d) is a simplified version of our method with SimCLR\cite{chen2020simple} framework, some details are not present.}
    \label{fig1}
\end{figure}

Since Dosovitskiy et al.,\cite{dosovitskiy2020image} first introduced self-attention mechanism in computer vision, Transformer, the most dominant architecture in natural language processing, has set records in various natural image classification tasks\cite{dai2021coatnet, zhai2022scaling, liu2022swin}.
Moreover, as Swin Transformer \cite{liu2021swin, liu2022swin} overcomes the limitation of high computational cost for dense pixel recognition with shifted local window,  this hierarchical Transformer is widely used for object detection and semantic segmentation. 
Meanwhile, since vision Transformer (ViT) usually has more parameters than the convolutional neural networks (CNN) counterpart, they are also more data-hungry. 
Thus, the superior performance of ViT depends on either transferring from models trained on large dataset\cite{dosovitskiy2020image} or pre-training on pretext task\cite{bao2021beit, he2022masked, xie2022simmim}.

Inspired by the success of ViT, medical image domain has also seen many works\cite{chen2021transunet, valanarasu2021medical, cao2021swin, hatamizadeh2022unetr, zhou2021nnformer, tang2022self} exploring transformer-based architecture, and most of them focus on semantic segmentation task.
Different from natural images, medical images present as a volume in many cases, such as ultrasound images, CT (computed tomography), and MRI (magnetic resonance imaging). 
The state of the arts on many public 3D medical image segmentation datasets are achieved by models using hierarchical Transformer backbone, like Swin UNETR\cite{tang2022self} and nnFormer \cite{zhou2021nnformer}.
Although existing works already beat classical CNN models only with new Transformer architecture,
we believe taking advantage of pretext task to pre-train Transformer can further boost the performance.

\begin{figure*}
    \centering
    \includegraphics[width=\linewidth]{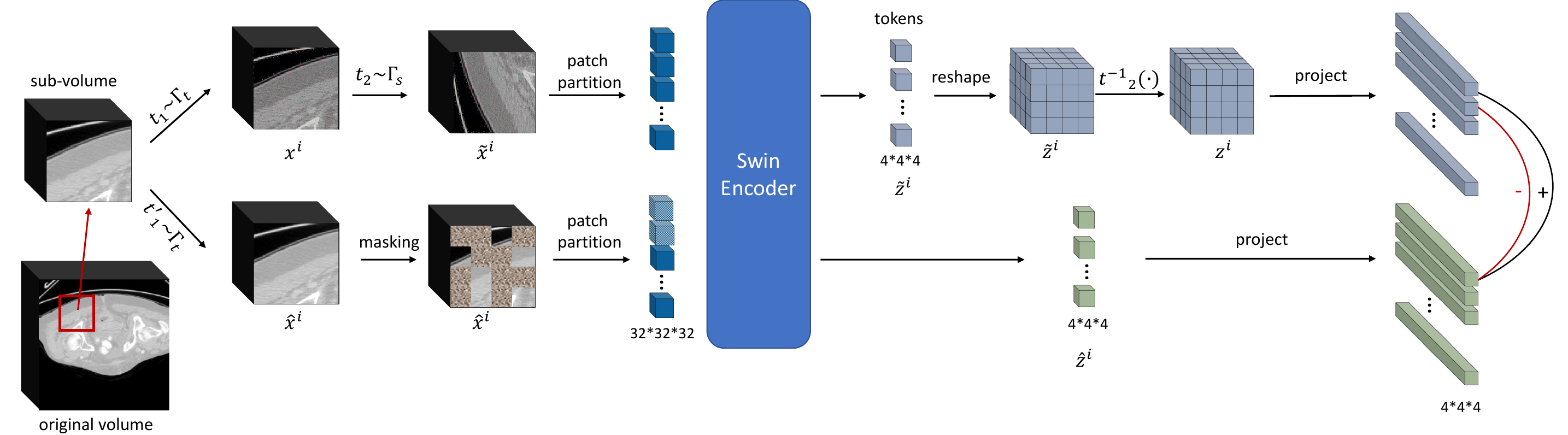}
    \caption{Detailed  workflow of \textbf{SimTROT}, our proposed token-wise representation embedded in a SimCLR framework. $\Gamma_{t}$ and $\Gamma_{s}$ represent random texture transformations and random spatial transformations (rotation and flip).  A Swin encoder projects different views $\Tilde{x}^{i}$ and $\hat{x}^{i}$ into 3D feature maps. Each cell in the cube is the feature representation of a group of neighboring input patches. A contrastive loss is then employed to learn useful features for the downstream task. }
    \label{fig2}
\end{figure*}

Borrowing self-supervised learning methods\cite{chen2020simple, he2020momentum, chen2020improved, grill2020bootstrap, chen2021exploring} from natural images is the first thought. 
SimCLR\cite{chen2020simple} and MoCO \cite{he2020momentum, chen2020improved} are two most popular representation learning techniques, but they both need a large number of samples to formulate enough negative pairs, either in a batch or in a memory buffer.
Although BYOL\cite{grill2020bootstrap} is not so sensitive to batch size, they still require batch size of at least 32.
However, in the case of 3D medical images, the batch size cannot be set as high as it can for 2D images because the memory footprint of a single 64×128×128 image is nearly equivalent to that of 64 128×128 2D images. 
\textbf{This significantly smaller batch size can reduce the effectiveness of these methods in learning 3D representations}.
On the other hand,
there is no large-scale segmentation dataset available because every volume containing hundreds of 2D slices makes voxel-wise annotation time-consuming.
The privacy issue also limits the quantity of unlabeled data accessible to the public.

Besides, there is another track of methods\cite{bao2021beit, he2022masked, xie2022simmim} specifically designed for ViT.
Among these, BEIT\cite{bao2021beit} uses an additional discrete VAE to get a codebook for each patch in the input images, and makes the transformer to predict the code of masked patches. 
However, Swin Transformer block will lead to the conflict between number of patches in the output and the number of visual tokens in the codebook, as is showed in Fig~\ref{fig1}.
For MAE\cite{he2022masked}, they firstly remove random patches of input images and feed the rest of patches into an encoder.
Then they append a slimmer transformer decoder to reconstruct the missing patches in the pixel space.
The problem with MAE when applied to hierarchical Transformer is that the missing patches in the input make the window partition very complex. 
SimMIM\cite{xie2022simmim} is quite similar with MAE, but they keep the masked patches. 
After getting output patch embeddings of the encoder, they deploy a fully connected layer to project feature maps back to input dimensions.
SimMIM can be directly used as a pre-training task for the Swin encoder ,
but effectiveness of SimMIM remains uncertain for medical images.
Also, given volumetric inputs, the dimension reduction is a cubic function and a single fully connected layer is not powerful enough to accurately restore visual information from patch embeddings.  

Considering the above limitations, this work presents a simple yet effective pre-training scheme to learn token-wise visual representation specifically for ViT.
In detail, we apply random combinations of texture augmentations on the input volumes to generate two different views, and feed them into a Swin encoder. 
After getting these two sequences of tokens, we maximise the agreement between the patch embeddings at the same position from two views.
In this way, given only one input volume, we can have a large batch of features, which satisfies the requirement of contrastive loss.
Meanwhile, being aware of effectiveness of mask modeling, we also add random masking on one view of the input volume as additional data augmentation, making the representation learning task more challenging.

However, during experiment, we find the pre-training process can easily converge to a trivial solution, where \textbf{the encoder simply projects the patches at the same position across different volumes into a constant feature vector}.
To address this problem, we design a ``rotate-and-restore" in our framework, as illustrated in Fig~\ref{fig2}. 
We deploy random rotation and flip on the view without masking, and later restore the position of the corresponding 3D feature maps.
As the order of tokens is mixed before the encoder, the model no longer simply clusters features with the same positional encoding.
This simple schemes can be applied to both SimCLR and BYOL frameworks, dubbed as SimPROT and B-TROT, in which, \textbf{TROT} stands for \textbf{\underline{T}}oken-wise representation learning with ``\textbf{\underline{ROT}}ate-and-restore".
In addition, for SimTROT, we add a weight on the term in the denominator that stands for similarity of feature tokens at the same position from different volumes to further address the collapse. 

To evaluate our method, we append an decoder to the Swin encoder after the pre-training, and then fine-tuning the whole classical downsampling-upsampling architecture with desne voxel-wise labels. 
We use two public medical image segmentation datasets, including Synapse for multi-organ CT segmentation and Brain Tumor Segmentation challenges.
We compare our methods with other pretext task on these two datasets regarding dice score and Hausdorff Distance 95\% (HD95).
Moreover, we use three different popular 3D Transformer models in the field of medical image, which are UNETR\cite{hatamizadeh2022unetr}. Swin UNETR\cite{tang2022self}, and nnFormer\cite{zhou2021nnformer}.
Experiments show that our method can be generalized to different architecture and boost their performance, and pre-training nnFormer with our devised pretext task surpasses current state of the arts on both tasks.
We also conduct comprehensive ablation studies on the key components in our proposed framework, to show motivation of each design. 

\section{Related Works}
\label{sec:related}
\textbf{Pre-training for vision transformer}
MAE\cite{he2022masked} and BEIT\cite{bao2021beit} are the first two works proposing pre-training tasks for ViT. The main difference between these two works is which space the mask modeling is based on.
MAE restores masked patches in the pixel space, and its variants include \cite{fang2022corrupted, chen2022context}, while BEIT along with its followers \cite{wei2022masked, zhou2021ibot, dong2021peco} predict masked patches in feature space. 
MSN\cite{assran2022masked} deploys a Siamese framework for mask modeling in Transfoemer.
Compared with our method, MSN projects the input volume into a global representation while we focus on patch-wise representation.
In addition, they remove masked patches same as MAE and then the input is not feasible for hierarchical ViT.
In addition, \cite{yun2022patch} also works on teaching ViT patch-level representation, but their patch-level self-supervision might not be suitable for Swin Transformer, as adjacent patches could contain very distinct semantic information after patch merging. 
On the other hand, some works apply the above pre-training approaches to domain specific tasks, like video segmentation\cite{wang2022bevt} and 3D point clouds\cite{yu2022point}

\textbf{Pre-training in medical image segmentation} To take advantage of unlabeled data, early works utilized  various pretext tasks to learn robust representations, like position prediction\cite{bai2019self, nguyen2020self}, Jigsaw puzzle\cite{li2020self, taleb2021multimodal, zhuang2019self}, image inpainting reconstruction\cite{chen2019self,zhou2021models, haghighi2021transferable}.
Later, with the emergence of self-supervised learning framework, some works\cite{chaitanya2020contrastive, zeng2021positional,hu2021semi, xie2020pgl, peng2021self} introduce this powerful pre-training to medical image  applications, and they redefine positive pairs and negative pairs based on domain specific knowledge. For example, Chaitanya et al\cite{chaitanya2020contrastive} noticed that within a 3D scan, adjacent images share similar anatomical objects and structures, and then their representations should be clustered in the feature space.
On the other hand, even though ViT attracts more attention in medical images analysis, there are not many papers related to pre-training the ViT.
\cite{tang2022self} is the only one we can find, but their main focus is a new invented architecture.
Their pre-training loss is a simple combination of rotation prediction, inpainiting reconstruction, and contrastive loss.
Therefore, the missing of pre-training methods designed specifically for medical image segmentation Transformer motivates this work.

\section{Methods}
\label{sec:methods}

\subsection{Model Architecture}

Due to the dramatic difference between natural images and medical images, segmentation models used in these two domains are quite different, especially considering that there are literally no 3D semantic segmentation tasks for natural images.
In the field of medical image segmentation, most of the emerging works that take advantage of ViT are  still based on classical U-net\cite{ronneberger2015u} structure, which consists of a downsampling path and an upsampling path with bypass connecting features in same stage of the two paths.
We add a multi-layer perceptron (MLP) on the downsapling path to act as en encoder, projecting input volumes into feature space. 
The proposed method mainly targets for the encoder, and after the pre-training finishes, we append the remaining decoder and bypass for the downstream fine-tuning. 

Although this work does not focus on improvement of architecture, for convenience of readers, we will briefly introduce three state-of-the-art 3D Transformer models used in this work.
(1) \textbf{UNETR} replaces CNN blocks in U-net with multi-head self-attention blocks in the downsampling path, and keeps the CNN decoder.   Given a input volume of size $128 \times 128 \times 128$, due to the limit of computational resource, they set the patch size to be 16 and the encoder outputs $8\times8\times8$ patch embeddings.
(2) \textbf{Swin UNETR} upgrades Transformer block in UNETR to Swin Transformer block. 
In our setting, we have four stages in the encoder, and each stage has several successive Swin Transformer blocks, including W-MSA and SW-MSA, same as in \cite{liu2021swin}.
As the attention is calculated within a local window, Swin Transformer block provides with a smaller patch size in the input, that is 4.
To produce a hierarchical representation, at the end of every stage, a patch merging layer merges group of $2\times 2\times 2$ adjacent patches. 
Therefore, there are $32\times 32 \times 32$ patches in the input while the output of encoder only gets $4\times4 \times 4$ patches.
(3) \textbf{nnFormer} uses Swin Transformer blocks for both encoder and decoder. 
Moreover, they make an adjustment on the patch merging layer. 
Instead of applying a linear layer on the concatenated neighbouring features, nnFormer deploys a convolutional layer with stride equal to 2 on the whole patch feature maps. 
The encoder of nnFormer outputs the same dimension of patches as Swin UNETR. 

\subsection{Pach-wise Representation Learning}
Fig~\ref{fig2} presents our method based on a SimCLR framework, and our method can also be fused into other representation learning framework, like BYOL.

\textbf{Combined with SimCLR}. Given an input volume, we firstly apply a random combination of texture transformations, $\Gamma_{t}(\cdot)$, including Gaussian noise, Gibbs noise, intensity scaling, and intensity shift.
Then, we get two different augmented views, $x^{i}$ and $\hat{x}^{i}$.
Only modifying voxel values is not sufficient to generate two views with strong contrast, and model would not learn meaningful representation.
In addition, we practice random masking on $\Tilde{x}^{i}$ that is used in \cite{bao2021beit} to enhance the difference between the two augmentations.
As shown in Fig~\ref{fig2}, we also apply random spatial transformations, $\Gamma_{s}(\cdot)$, including random rotation and random flip,  on $x^{i}$ to get $\Tilde{x}^{i}$.
The benefits of spatial transformation are two-fold, the first one is making the comparison between $\Tilde{x}^{i}$ and $\hat{x}^{i}$ more significant, and the other one is avoiding trivial solution. 
The trivial solution issue will be discussed later in section \ref{sec3.3}.
After we get the two augmented volumes , we divide them into small patches to feed into the encoder $Enc_{\theta}(\cdot)$. 
The output 3D feature maps $\Tilde{z}^{i}, \hat{z}^{i} \in \mathbb{R}^{C*h*w*d}$ contain a sequence of patch embeddings of dimension C. $h*w*d$ is the patch dimension,
and it is the same as the input patch dimension for UNETR. 
For Swin Encoder, the dimension is reduced due to patch merging layer.
The next step is deploying a reverse spatial transformation on $\Tilde{z}^{i}$ to restore the order of patch embeddings back to original position $z^{i} = \Gamma^{-1}_{s}(\Tilde{z}^{i}) $. 
Thus, patch embeddings $z^{i}_{m}, \hat{z}^{i}_{m} \in \mathbb{R}^{c}$ that share the same index stores high-dimension information of the same patch or the same area in the input volume. 
Only then is maximizing the similarity between these representations reasonable.
The contrastive loss regarding positive pair $(z^{i}_{m},\hat{z}^{i}_{m}) $ is defined as:
\begin{equation}
\centering
    L_{con}(z^{i}_{m}, \hat{z}^{i}_{m}) = -log\frac{exp(sim(z^{i}_{m}, \hat{z}^{i}_{m}) / \tau )}{ \sum_{z\neq z^{i}_{m}}  exp(sim(z^{i}_{m}, z)/\tau) }
    \label{eqt1}
\end{equation}
where $sim(\cdot)$ is a function measuring similarity between two feature vectors, and in this work, we use cosine distance for $sim(\cdot)$. $\tau$ is the temperature parameter, and we set it as 0.5.

\textbf{Combined with BYOL} To fuse our method into BYOL (bootstrap your own latent), the original $Enc_{\theta}(\cdot)$ is used for the online path, and we need to introduce another encoder $Enc_{\xi}(\cdot)$ with the same structure but different weight $\xi$ for the target path. 
The data pre-processing flow is the same as showed in Fig~\ref{fig2}, and when we get the two augmented views $\Tilde{x}^{i}, \hat{x}^{i}$, we input them into the above two encoders to generate feature maps, $\Tilde{z}^{i} = Enc_{\theta}(\Tilde{x}^{i}),  \hat{z}^{i} = Enc_{\xi}(\hat{x}^{i})$.
The difference between our method and BYOL is that the output of encoder is one-dimension feature vector, while we do not apply global average pooling and keep embeddings for every patches. 
Moreover, following settings in BOYL, additional projection heads $g_{\theta}(\cdot), g_{\xi}(\cdot)$are added on the two paths to output patch-wise projections, respectively.
For online path, the last stage outputs a prediction with a head $pred(\cdot)$, making a asymmetric structure between the online and target pipeline.
After normalizing every patch embeddings,  we define the following mean squared error between the predictions and target projections,

\begin{equation}
\centering
    L_{byol}(z^{i}_{m}, \hat{z}^{i}_{m}) = \left\|pred( g_{\theta}( z^{i}_{m})) -g_{\xi}(\hat{z}^{i}_{m}) \right\|^{2}
    \label{eqt2}
\end{equation}
The gradient of $L_{byol}$ is only back-propagated to parameters $\theta$ in the online path, and $\xi$ is updated by an exponential moving average of $\theta$. 

\begin{figure}
    \centering
    \includegraphics[width=\linewidth]{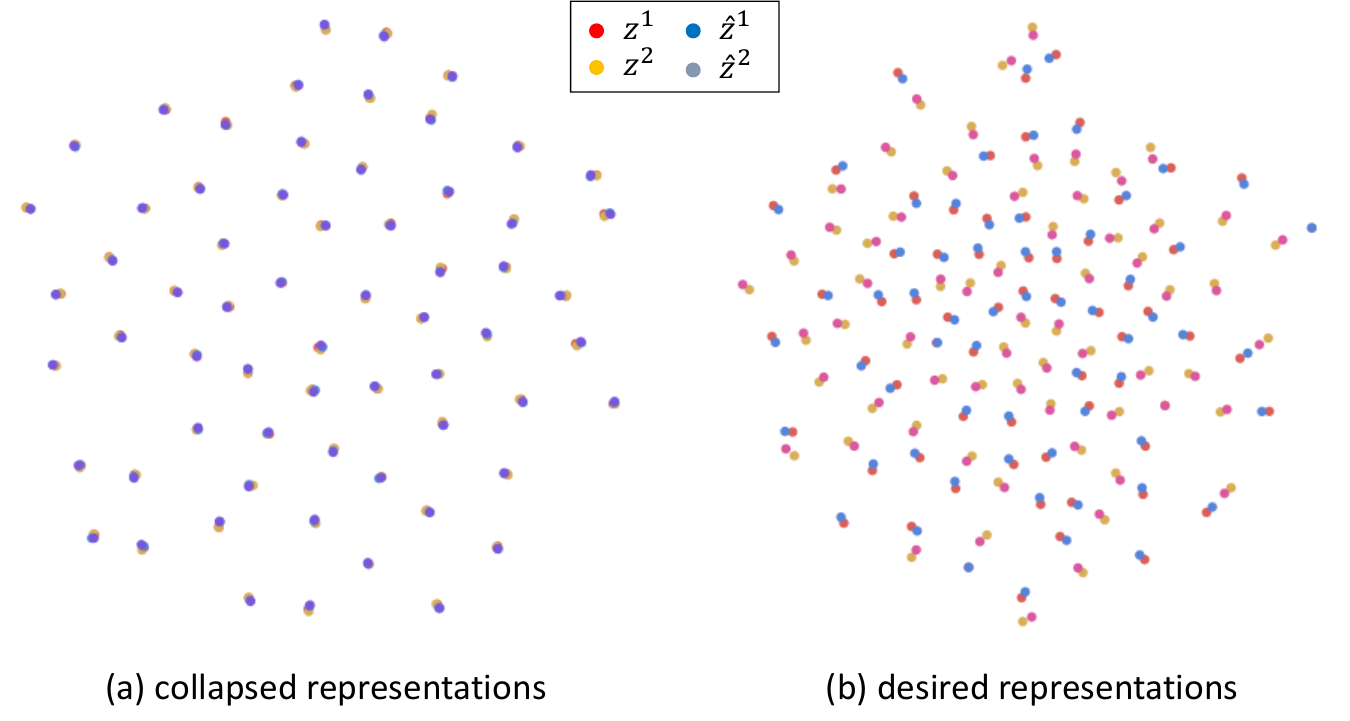}
    \caption{Visualization of representation collapse in the feature space. There are four different feature maps, and each feature maps contains $4*4*4$ embeddings. (a) displays a pattern where all four feature maps are exactly the same and embeddings at the same coordinates are gathered together. In contrast, the ideal learned representations should be like (b), where only embeddings of different views from the same volume are clustered based on position in the feature map, and others are separated. Best viewed in color.}
    \label{fig3}
\end{figure}

\subsection{Tackling Representation Collapse}
\label{sec3.3}
To learn useful representation, self-supervised learning works train the model to predict different augmented views from one other.
This scheme can lead to a trivial solution where the model simply encodes all input into the same features, which is called representation collapse.
SimCLR tackles this problem by introducing negative pairs so that the prediction problem not only reduces the distance between representations of two augmentations from the same image, but also discriminates features belonging to different images.
Meanwhile, BYOL circumvents the trivial solution by constructing two asymmetric paths for different views, and the fact that gradient of the loss is not backpropagated to one path prevents the two paths learning the exact same features.
However, in the context of patch-wise representation learning, the training can end up with collapsing in a new way. 
For instance, given feature maps of two different views, although $z^{i}_{m}$ is prevented from quickly converging to $z^{i}_{n}$,
 the model can still arrive at a trivial solution where patch embedding at the same position but from different volumes are constantly close in the latent space, that is $z^{i}_{m}$ being the same as $z^{j}_{m}$.
Regarding SimCLR, although the term $z^{i}_{m}*z^{j}_{m}$ will be present in the denominator of equation(\ref{eqt1}) when the batch size is larger than 1, the large size of other terms in the denominator can diminish the effect of similarity  between $(z^{i}_{m},z^{j}_{m})$.
In terms of BYOL, there is also no penalization mechanism preventing the above collapse. 
What makes it worse is that positional embeddings are the same across volumes. 
The positional embeddings can inform the encoder of a given patch's position so that patch-wise representations may be discriminated just by their positions.

This motivates the use of ``rotate-and-restore" in our pipeline. 
As is described in section 3.2, the rotation and flip can change the order of patches after the patch partition layer.
In equation(\ref{eqt1}) and equation(\ref{eqt2}), the losses no longer predict representation from the other at the same position in the input volume.
With the influence of positional embedding, it is hard for model to just remember positions , instead the model is forced to learn meaningful representation extracted from visual information in each patch during the task.
Talking about spatial transformation itself, in natural language processing, the input is a 1-D sequence of words, and change in order of words or characters within each word only results in non-sense sentences. 
While for vision tasks, regardless of the rotation or the flip exerted on the input images, it always leads to a different view but with the same visual meaning.
On the other hand, the second strategy we use to avoid collapsing is quite straightforward, that is adding a weight on the term that represents similarity between the same patches from different volumes in the denominator. 
Thus, the loss will penalize more for the trivial solution. 
This method is only applicable for SimCLR framework, since the loss in BYOL does not have such terms.
The new loss function with the attention weight is defined as

\begin{equation}
\centering
\footnotesize
\begin{split}
\centering
     L_{wcon}(z^{i}_{m},\hat{z}^{i}_{m}) = & -z^{i}_{m}\cdot\hat{z}^{i}_{m} / \tau  + 
     log (\sum_{z \notin z_{m}}  exp(z^{i}_{m} \cdot z)/\tau \\
     & + w* \sum_{j\neq i} exp(z^{i}_{m}\cdot z^{j}_{m})/\tau)
     \label{eqt3}
\end{split}
\end{equation}
where $w$ is the attention weight. 
The default value of $w$ in this work is set as 5.
If the weight is too small, the value of other terms in the denominator will weaken the impact of $z^{i}_{m}$ and $z^{j}_{m}$ being the same.
If the weight is too large, (\ref{eqt3}) will ignore the numerator and fail to reduce the gap in latent space between different views from the same input. The quantitative results on impact of $w$ can be seen in section~\ref{sec:ablation} of ablation studies.


\newcolumntype{Y}{>{\centering\arraybackslash}X}
\begin{table*}[t]
    \footnotesize
    \centering
    \caption{Comparisons between state-of-the-art methods and the proposed methods w.r.t. segmentation dice scores on multi-organ segmentation dataset (Synapse). The last three rows are our proposed methods, and SimPROT-W means SimPROT pre-training with attention weight $w=5$ in the loss function.}
    \label{table1}
    \begin{tabularx}{\linewidth}{cc|YYYYYYYYY}
    \toprule
\multirow{2}{*}{Backbone}  &\multicolumn{1}{c|}{\multirow{2}{*}{Pretrain}} & \multicolumn{9}{c}{Dice$\%$ $\uparrow$}  \\ \cline{3-11}
 &&\multicolumn{1}{c}{spleen} & \multicolumn{1}{c}{RK} & \multicolumn{1}{c}{LK} & \multicolumn{1}{c}{GB}  & \multicolumn{1}{c}{liver} & \multicolumn{1}{c}{stomach}  & \multicolumn{1}{c}{aorta} & \multicolumn{1}{c}{pancreas} & \multicolumn{1}{c}{\textbf{avg}} \\
 \hline
 UNETR  &N/A &83.85 &81.62 &81.45 &55.10 & 94.56 & 69.37 & 87.21 & 57.47 &76.33 \\
UNETR  &MAE &84.61  &82.94  &83.62  &55.42    &93.09  &68.45    &87.63  &57.48 &76.64 \\ \hline
Swin UNETR  &N/A  &88.26 &84.22 &85.91 &61.63 &94.81 & 77.94 & 91.49 & 67.78 & 81.51 \\
Swin UNETR  &RCR &88.57  &85.44  &86.13  &62.77    &95.42  &78.12   &91.58  &68.84 &82.10 \\
\hline
nnFormer  & N/A &86.98  &84.21  &86.04  &65.22    &95.30  &82.85    &90.36  &82.77 &84.21 \\
nnFormer  & SimCLR &88.66  &86.38  &85.58  &67.19    &96.73  &83.90    &91.41  &81.83 &85.21 \\
nnFormer  & BYOL  &88.50 &86.65 &85.70 &68.60 &96.23 &80.74 &92.57 &79.43 &84.99 \\
nnFormer  & Simmim &89.94  &87.49  &83.01  &69.50    &96.60  &84.87    &92.45  &81.33 &85.63 \\
nnFormer  &RCR &88.27  &86.43  &85.87  &68.98    &96.46  &83.01   &91.13  &80.92 &85.14 \\
\hline
nnFormer  & SimPROT &90.27  &86.21  &86.44  &68.17    &$\textbf{96.75}$  &$\textbf{86.66}$  &92.05  &81.76 &86.03 \\
nnFormer  & BPROT &88.63  &$\textbf{86.92}$  &86.81  &71.97    &96.10  &84.17  &$\textbf{92.47}$  &79.87 &85.87 \\
nnFormer  & SimPROT-W &$\textbf{91.30}$  &86.65  &$\textbf{87.16}$  &$\textbf{72.31}$   &96.55  &86.62    &90.95  &$\textbf{82.80}$ &$\textbf{86.79}$ \\
\bottomrule
\end{tabularx}
    
\end{table*}

\begin{table*}[t]
    \footnotesize
    \centering
    \caption{Comparisons between state-of-the-art methods and the proposed methods w.r.t. segmentation dice scores as well as HD95 on Brain Tumor segmentation dataset. }
    \label{table2}
    \begin{tabularx}{\linewidth}{cc|YYYY|YYYY}
    \toprule
\multirow{2}{*}{Backbone}  &\multicolumn{1}{c|}{\multirow{2}{*}{Pretrain}} & \multicolumn{4}{c|}{Dice$\%$ $\uparrow$}  & \multicolumn{4}{c}{HD95 $\downarrow$} \\ \cline{3-10}

 &&\multicolumn{1}{c}{ET} & \multicolumn{1}{c}{TC} & \multicolumn{1}{c}{WT} & \multicolumn{1}{c|}{\textbf{avg}} &\multicolumn{1}{c}{ET} & \multicolumn{1}{c}{TC} & \multicolumn{1}{c}{WT} & \multicolumn{1}{c}{\textbf{avg}} \\
 \hline
UNETR  &N/A   &77.34  &83.40  &90.50 &83.75    &8.72  &6.08   &4.64  &6.48  \\
 
UNETR  &MAE &78.09  &83.64  &90.10  &84.24    &5.16  &6.77   &3.97  &5.30  \\
 \hline
Swin UNETR  &N/A  &78.98  &84.12  &91.19  &84.77    &4.69  &6.77   &3.97  &5.14  \\

Swin UNETR  &RCR &79.74  &84.64  &91.35  &85.24 &4.72    &4.93  &4.44   &4.69 \\

\hline
nnFormer  & N/A &81.18  &85.21  &90.56  &85.65    &3.95  &6.33   &3.89  &4.73 \\

nnFormer  & SimCLR &80.09  &85.41  &91.00  &85.50    &4.07  &4.88   &4.10  &4.35   \\

nnFormer  & BYOL &81.48  &85.33  &90.87  &85.89   &4.69   &4.65  &3.91 &4.42 \\
nnFormer  & Simmim &81.91  &85.21  &90.56  &85.91    &4.86  &5.19   &4.14  &4.73 \\

nnFormer  & RCR &81.19  &85.38  &90.93  &85.83    &5.67  &4.77   &3.67 &4.70 \\

\hline
nnFormer  & SimPROT &81.35  &85.57  &\textbf{91.43}  &86.12     &4.09   &\textbf{4.19}  &\textbf{3.53} &3.94 \\
nnFormer  & BPROT &81.25  &\textbf{86.10}  &90.86  &86.07    &3.95  &4.67   &3.84  &4.15  \\
nnFormer  &SimPROT-W &\textbf{81.98}  &85.82  &91.12  &\textbf{86.29}     &\textbf{2.71}   &4.51  &3.73 &\textbf{3.65}  \\

\bottomrule
\end{tabularx}
    
\end{table*}

\section{Experiment}
\subsection{Dataset}
\textbf{Synapse Multi-organ Segmentation} 
Synapse is from the \emph{Multi-Atlas Labeling Beyond the Cranial Vault Challenge}, containing 30 volumes of CT scans. 
For a fair comparison, we use the same setting in ~\cite{chen2021transunet}. 18 cases are used for training, and the rest 12 cases are used for testing.
In terms of evaluation metrics, we calculate both the dice score and  Hausdorff Distance 95\% (HD95) between prediction mask and the label on test set.
The organs to be segmented include spleen, right kidney(RK), left kidney(LK), gallbladder(GB), liver, stomach, aorta, and pancreas, 8 different classes in total.

\textbf{Brain Tumor Segmentation}
This dataset is one of the segmentation tasks from Medical Segmentation Decathlon dataset\cite{antonelli2022medical}, and the task contains 484 MRI images with annotations of tumor regions. 
For every MRI, there are four channels representing different MR modalities, which are FLAIR, T1w, T1gd, and T2w. 
In experiment, we utilize all of the four modalities and the input can be viewed as a four-channel volume.
We follow the data split in UNETR\cite{hatamizadeh2022unetr}, dividing the 484 images into three parts, train/validation/test, with ratio $80\%/15\%/5\%$.
For evaluation, we report the dice sore and HD95 of three types of tumor, namely the whole tumor (WT), enhancing tumor (ET), and tumor core (TC).

\textbf{Pre-processing} 
We also follow most of the pre-processing steps in \cite{hatamizadeh2022unetr}.
  For Synapse, the sub-volume size is [64, 128, 128], and it is [128, 128, 128] for brain tumor task. The batch size for both two datasets is set as 2 during pre-training and fine-tuning.

\begin{figure*}
    \centering
    \includegraphics[width=0.9\linewidth]{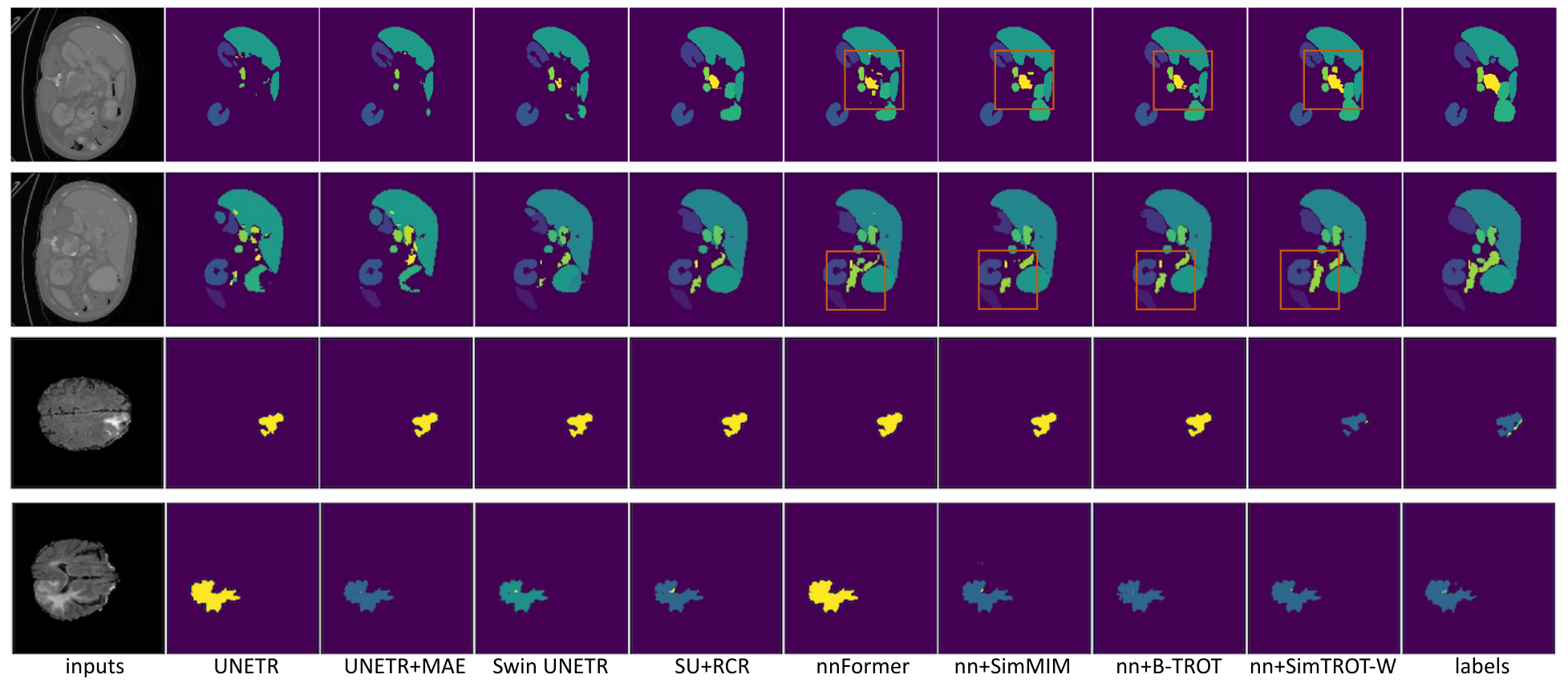}
    \caption{Visualization of segmentation results of different methods on Synapse and Brain Tumor dataset. The above two rows are results from multi-organ segmentation, and the two rows at the bottom are from tumor segmentation task.  ``SU" is short for Swin UNETR, and ``nn" is short for nnFormer. Best viewed in color. }
    \label{fig4}
\end{figure*}

\subsection{Experiment Settings}

\textbf{Model structure}. This work uses three model architectures, UNETR, Swin UNETR, and nnFormer. For each model, we try to keep all the structure hyperparameters the same as in their original github repositories. 
All of the models have 4 resolution stages.
Since UNETR is not a hierarchical structure, number of patches as well as patch embedding dimension is kept the same all the way downside in the encoder. 
For Swin UNETR and nnFormer, to get next stage, the patch dimension will be reduced by 2 for all three axis and the embedding dimension will double. 
For Synapse dataset, the patch size in the patch partition layer is [8, 16, 16] in UNETR, [1, 2, 2] in Swin UNETR, and [2, 4, 4] in nnFormer.
As for brain tumor dataset, the patch size in the patch partition layer is [16, 16, 16] in UNETR, [2, 2, 2] in Swin UNETR, and [4, 4, 4] in nnFormer.
When the batch size is set as 2, all models can successfully run on a single NVIDIA 1080TI gpu with 10G memory on both datasets under the above settings.

\textbf{Optimizer}. During pre-training, the default optimizer is SGD with learning rate being 0.0001 for both two datasets, and we also enable Nesterov momentum and set the momentum to be 0.99. The pre-training epoch for Synapse dataset is 1000 and the epoch for brain tumor dataset is 500. For fine-tuning stage, we follow the setting in \cite{zhou2021nnformer}, using an SGD optimizer with learning rate being 0.01 and momentum being 0.99 on both multi-organ and brain tumor segmentation tasks. Also, a "poly" decay strategy is employed to gradually reduce learning rate after epochs of training. The number of iterations is set to be 1000.

\begin{table}[t]
     \small
    \centering
    \caption{Evaluation of our methods on three different segmentation Transformer models. Results on two datasets demonstrate our SimPROT-W results in better segmentation accuracy compared with training from scratch.}
    \label{table3}
    \resizebox{\columnwidth}{!}{
    \begin{tabular}{ccc|cc}
    \toprule
 Dataset & Model & Pretrain & Dice\% $\uparrow$ & HD95 $\downarrow$ \\
\hline
\multirow{6}{*}{Synapse}&UNETR &N/A &76.33 & 30.10\\
  &UNETR &SimPROT-W &77.98 &26.36 \\
 &Swin UNETR &N/A &81.51 &24.91 \\
 &Swin UNETR &SimPROT-W &82.61 &20.85 \\
  &nnFormer &N/A &84.21 &17.15 \\
 &nnFormer &SimPROT-W &\textbf{86.79} &\textbf{9.19} \\
\hline
\multirow{6}{*}{Brain Tumor}&UNETR &N/A &83.75 &6.48 \\
  &UNETR &SimPROT-W &84.38 &5.81 \\
 &Swin UNETR &N/A &84.77 &5.14 \\
 &Swin UNETR &SimPROT-W &85.29 &4.74 \\
  &nnFormer &N/A &85.65 &4.73 \\
 &nnFormer &SimPROT-W &\textbf{86.29} &\textbf{3.65} \\
\bottomrule
\end{tabular}
}
\end{table}

\subsection{Comparison with the State of the Arts}

To validate the effectiveness of our methods, we implement a series of pre-training methods and report their segmentation accuracy on the downstream segmentation task as comparison.
First of all, we include the performance of models trained from scratch, and these results are the reference on how much improvement the pre-training stage brings. 
We try our approaches on all models, and nnFormer outputs the best performance.
Thus, we list the results of nnFormer in Tables~\ref{table1} \&~\ref{table2}, and the results of our methods on other architectures can be seen in Table~\ref{table3}.
The three methods at the bottom of  Tables~\ref{table1} \&~\ref{table2} are our proposed methods.
One of the key observations we can make is that our method with nnFormer gets the superior performance than other pre-training methods. Fine-tuning with our pre-trained weights has the highest dice in all eight organs in Synapse dataset, and they achieve the highest dice score as well as lowest HD95 for all three different types of tumors in the Brain Tumor segmentation dataset

\textbf{Comparison with self-supervised learning methods}. We implement the two most popular self-supervised learning methods, SimCLR\cite{chen2020simple} and BYOL\cite{grill2020bootstrap}. In Synapse dataset, using SimCLR and BYOL increase dice scores of training from scratch by 1.00 and 0.78, respectively.
Although both of them provide with improved performance, there still exists gap between their results and our patch-wise representation learning methods. 
This implies that, when not given large batch size,  using patch-wise representation instead of global representation is a better alternative. 

\textbf{Comparison with pretext tasks for Transformer}.
Since MAE is not applicable to hierarchical architecture, we only try it on UNETR encoder and it turns out that the benefits of MAE is limited, e.g. increasing dice score from 76.33 to 76.64 in Table~\ref{table1}.
We choose SimMIM\cite{xie2022simmim} as the representative of pretext task to be applied to Swin Transformer, and results show that it is slightly better than self-supervised learning methods. 
\cite{tang2022self} is the only work designed for Transformer in medical image applications.
In that paper, they apply the RCR (rotation prediction + contrastive + inpainting reconstruction) loss on the encoder from Swin UNETR, thus we include the experiment of Swin UNETR + RCR.
Also, we test RCR on nnFormer. 
Notice that, for the pre-training, the experiment setting between RCR in \cite{tang2022self} and in this work is slightly different regarding the dataset.
RCR in \cite{tang2022self} uses external dataset besides the training data, while for fairness all the methods in this paper only use the data in training set within each task.
Results show that, without external dataset, RCR learns less useful representations than our methods do.

\textbf{Results for Limited labeled data}
We also test SimTROT on Synapse dataset given only part of the training data to test data-efficiency of our proposed method.
From Fig \ref{fig5}, we can see the gap between SimTROT and training from scratch is more significant when there are fewer labeled volumes. For example, for $10\%$, that is only one labeled volume, the average dice score of all organs are 22.05 and 10.80 for SimTROT and scratch, and the performance improvement is 11.25, better than the gap when using all training data, 2.58.
Also, when using only $70\%$ data, SimTROT achieves comparable segmentation accuracy as scratch with all training data, which means that SimTROT can reduce the cost of annotation by $30\%$ for Synapse dataset.

\begin{table}[t]
    \footnotesize
    \centering
    \caption{Ablation study of key components in the proposed framework. ``mask'' stands for mask modeling, $\Gamma_{s}$ stands for the ``rotate and restore'' mechanism, and ``weight'' means adding attention weight in the contrastive loss. The dataset used for this series of experiments is Synapse.}
    \label{table4}
    \resizebox{\columnwidth}{!}{
    \begin{tabular}{cccc|cc}
    \toprule
 Framework & mask & $\Gamma_{s}$  & weights & Dice$\%$ $\uparrow$ & HD95 $\downarrow$ \\
\hline
\multirow{5}{*}{SimCLR} & & & &85.01 &15.80 \\
  &$\checkmark$ & & &85.60 &11.51 \\
 & &$\checkmark$ & &85.56 &13.48 \\
 &$\checkmark$ &$\checkmark$ & &86.03 &14.01 \\
  &$\checkmark$ &$\checkmark$ &$\checkmark$ &$\textbf{86.79}$ &$\textbf{9.19}$  \\
 
\hline
\multirow{4}{*}{BYOL} & & & &84.82 & 15.28\\
  &$\checkmark$ & & &85.29 &14.13 \\
 & &$\checkmark$ & &85.35 &13.62 \\
  &$\checkmark$ &$\checkmark$ & &$\textbf{85.87}$ &$\textbf{11.80}$ \\
\bottomrule
\end{tabular}
}
\end{table}

\subsection{Ablation Studies}
\label{sec:ablation}
\textbf{Model architecture}. Table \ref{table3} studies performance of our methods with different model architectures. As can be seen, on both two datasets, the proposed pre-training task SimPROT-W can benefit all three Transformer-based models, and the improvements are more significant than MAE on UNETR as well as RCR on Swin UNETR.
We also notice that, with pre-training task, UNETR and Swin UNETR still fail to beat nnFormer trained from random initialization.
Although it shows that model architecture may play a more important role that pre-training methods in deciding segmentation accuracy, searching new model architecture requires intense trials with no clue,  
while it is likely that our proposed method can work well on new ``best'' model.   

\textbf{Design of our method}.
In Table \ref{table4}, we test various combinations of the three key components in our proposed framework, adding mask (mask), ``rotate and restore" ($\Gamma_{s}$), and attention weights (weights).
We conduct the ablation studies on both SimCLR framework and BYOL framework. 
The results prove that solely adding a mask or applying spatial transformation make SimCLR and BYOL learn better representations. 
This is because only texture augmentation like adding noise or changing intensity scale is not a strong augmentation, and models cannot learn robust features by contrasting two weak views.
Also, combining them both can bringing in further performance boost.
While attention weight is not applicable to loss function in BYOL, a suitable weight $w$ in SimCLR can improve the dice score by 0.76, which demonstrates the power of this simple strategy to avoid feature collapse.

\begin{figure}
    \centering
    \includegraphics[width=\linewidth]{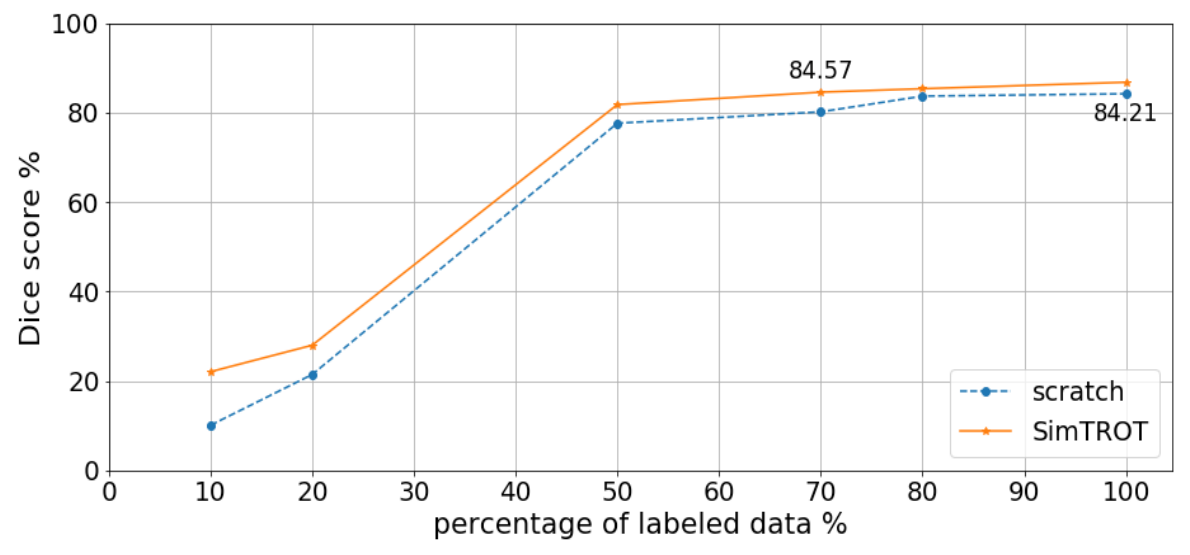}
    \caption{Performance of training from scratch and our SimTROT on Synapse dataset given different percentages of labeled training data. }
    \label{fig5}
\end{figure}

\textbf{Mask ratio and attention weight}. These two hyperparameters are also very important to our defined pretext task. To evaluate their impact, we try different values of mask ratio applied in the augmentation and attention weight $w$ added to contrastive loss in the SimCLR framework on Synapse dataset. 
When changing mask ratio, we keep the attention weight as 5. 
In the context of this work, larger mask ratio means more difficult contrast task between different views. 
Meanwhile, the masked patches cannot account for over 85\% of total patches, as there is no visual information left for the model. 
In terms of attention weight $w$, we also observe its high impact  on performance. 
The default mask ratio for attention weight is 0.75.
Same as the analysis in Section 3.3, neither small $w$ nor large $w$ is suitable. Through experiments, we find that setting $w=5$ gives the best downstream performance.
Actually, this hyperparameter is also relevant to the number of terms in the denominator of $L_{wcon}$(\ref{eqt3}), decided by batch size and the number of tokens in the output feature maps. 
We suggest trying various values of $w$ when used in settings different from this work.

\begin{figure}
    \centering
    \includegraphics[width=\linewidth]{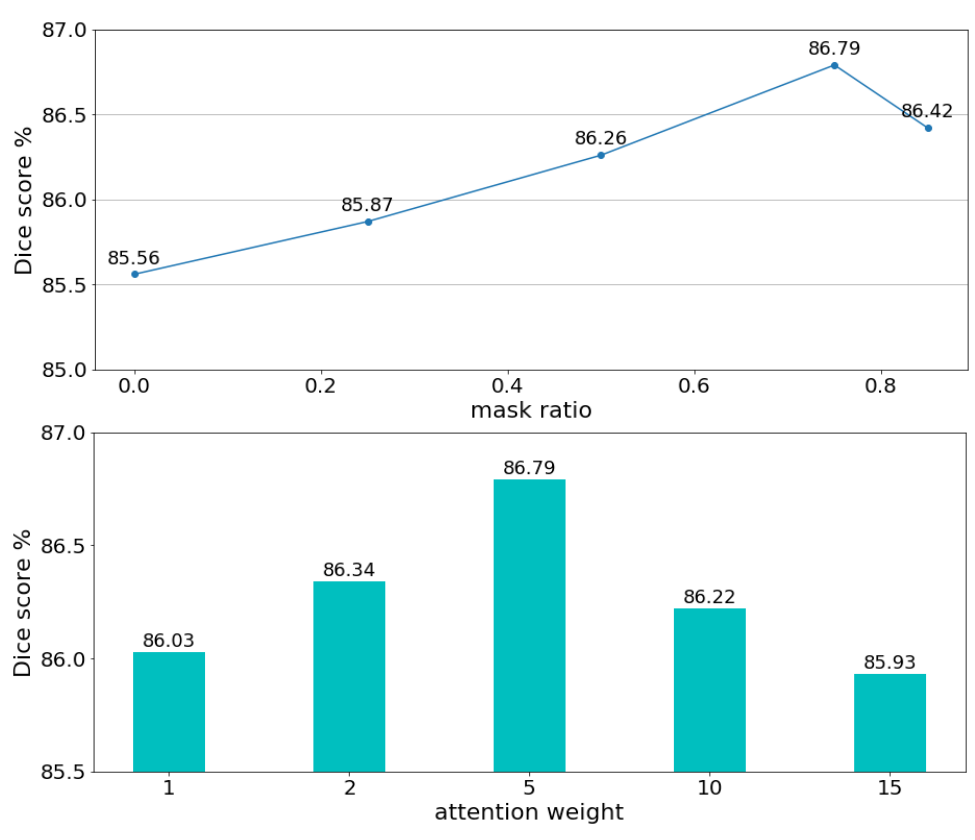}
    \caption{How \textbf{mask ratio} in the augmentation stage and \textbf{attention weight} $w$ in $L_{wcon}$ impact the dice score in the fine-tuning. The above results show that generally large mask ratio learns better representation than small mask ratio, but cannot exceed 0.75. When comparing different values of $w$, we find setting $w$ = 5 is the best option accordingly.}
    \label{fig5}
\end{figure}

\section{Conclusion}
Without access to large dataset, pre-training with a pretext task is then the key to success of ViT on computer vision tasks.
This work further proves this argument in the field of medical image analysis, and focuses on a more challenging task, volumetric semantic segmentation.
Considering the limitation of biomedical data and characteristics of segmentation task, we propose to predict token-wise representations from different augmented views of the same volume, rather than working on a single global representation for each augmentation.
Moreover, we devise two strategies to address the issue of representation collapse: one is ``rotate-and-restore", and the other is modifying the denominator in contrastive loss.
This simple yet effective method can be used with different Transformer models, either hierarchical or not, and with different self-supervised learning architectures. 
nnFormer pre-trained with our SimTROT-W demonstrates the best performance on both multi-organ segmentation and brain tumor segmentation tasks, when compared with other pretext approaches.
{\small
\bibliographystyle{ieee_fullname}
\bibliography{11_references}
}

\ifarxiv \clearpage \appendix
\label{sec:appendix}

 \fi

\end{document}


\title{\paperTitle \\ Supplemental Material}
\author{\authorBlock}
\maketitle

\appendix
\label{sec:appendix}


{\small
\bibliographystyle{ieee_fullname}
\bibliography{11_references}
}